\documentclass[11pt, a4paper]{article}

\usepackage[a4paper, total={16cm, 25cm}]{geometry}
\usepackage{amsmath}
\usepackage{amssymb}
\usepackage{graphicx}
\usepackage{natbib}
\usepackage{amsthm}
\usepackage{color}
\usepackage{multirow}
\usepackage{url}
\usepackage{authblk}
\usepackage{footmisc}

\usepackage{tikz}
\usetikzlibrary{positioning, arrows.meta, shapes.geometric, calc}

\newtheorem{rational for conjecture}{Rational for Conjecture}

\usepackage{algorithm}
\usepackage{algorithmic}

\DeclareMathOperator*{\argmin}{arg\,min}

\title{Improved Predictive Performance and Interpretability for Mesomorphic Neural Networks Using Local Fidelity Regularization}

\author[1,2]{Hugo L. Hammer}
\author[2]{Vajira Thambawita}
\author[1]{Kristoffer Herland Hellton}
\author[2,1]{Pål Halvorsen}

\affil[1]{Oslo Metropolitan University (OsloMet)}
\affil[2]{Simula Metropolitan Center for Digital Engineering (SimulaMet)} 

\begin{document}

\maketitle

\begin{abstract}
Interpretable Mesomorphic Neural Networks (IMNs) offer a promising framework that combines the predictive power of deep neural networks with the interpretability of linear models. However, the original formulation lacks safeguards to ensure that the learned interpretations are in fact reliable. In particular, the network is free to concentrate all explanatory variance into a single weight of the linear output layer, achieving strong predictive performance while producing interpretations that are largely meaningless. Paradoxically, the L1 penalty proposed to encourage sparse solutions exacerbates this problem by further incentivizing such degenerate configurations.

To address this vulnerability, we introduce Local Fidelity Regularization (LFR), a novel penalty term that prevents degenerate weight collapse by aligning the linear output weights with local data variations. This structural constraint guarantees faithful explanations and substantially improves the reliability of model interpretations. Furthermore, empirical evaluations across the OpenML benchmark suite demonstrate that LFR does not compromise accuracy for explainability; rather, it achieved improved AUROC over the unregularized IMN. By yielding results highly competitive with state-of-the-art black-box models, LFR provides the dual benefit of reliable interpretability and superior predictive performance. Source code and usage instructions are available at \url{https://github.com/hugohammer/LFR-IMN.git}.
\end{abstract}

\section{Introduction}

Tabular data is arguably the most abundant and commonly encountered data format within statistics and machine learning. This structured format serves as the foundational backbone for critical applications across diverse domains ranging from Electronic Health Records (EHRs) in medicine~\cite{cowie2017electronic} to credit scoring profiles in finance~\cite{dastile2020statistical} and engineering and material property databases in manufacturing~\cite{horton2025accelerated}.

While deep learning architectures, such as Convolutional Neural Networks (CNNs), have revolutionized the analysis of unstructured data like images by exploiting local spatial correlations, tabular data presents a fundamentally different challenge. Tabular datasets are inherently heterogeneous, lack a spatial or sequential structure, and often require modelling sharp, discontinuous decision boundaries. Due to this lack of suitable inductive bias in standard neural networks, they have historically struggled to achieve state-of-the-art predictive performance on tabular tasks. Instead, traditional statistical methods like Support Vector Machines (SVM) and tree-based ensembles, particularly Random Forests and Gradient Boosting, have remained the dominant standard, as their splitting mechanisms naturally accommodate the complex, discrete nature of tabular features~\cite{grinsztajn2022tree}.

Recently, however, deep learning architectures explicitly tailored for tabular data have emerged to bridge this performance gap. Models such as TabNet~\cite{arik2021tabnet} have introduced sequential attention mechanisms that perform instance-wise feature selection, effectively mimicking the sparse, discrete decision-making processes of tree ensembles within a differentiable framework. Concurrently, architectures like TabResNet~\cite{kadra2021well} have successfully adapted residual connections to tabular inputs. By mitigating the vanishing gradient problem and preventing overfitting on noisy features, these residual networks allow for deeper, more expressive representations of heterogeneous data. Furthermore, Neural Oblivious Decision Ensembles (NODE)~\cite{PopovMB20} explicitly fuse the strengths of both paradigms by embedding differentiable oblivious decision trees within a deep neural network, enabling the architecture to natively learn sharp, hierarchical boundaries. Finally, the Feature Tokenizer Transformer (FT-Transformer)~\cite{gorishniy2021revisiting} extends the highly successful Transformer architecture to the tabular domain by tokenizing continuous and categorical inputs into embeddings and using self-attention to capture complex feature interactions. Together, these advancements have demonstrated that with the correct architectural inductive biases, neural networks can achieve highly competitive performance to state of the art methods on tabular tasks.

A significant limitation of deep learning models for tabular data, such as TabResNet and standard multilayer perceptrons (MLPs), is their inherent lack of interpretability, which remains a strict prerequisite for safety-critical applications. Recently, the Interpretable Mesomorphic Neural Network (IMN)~\cite{kadra2024interpretable} was proposed to bridge the gap between interpretability and state-of-the-art predictive performance. The IMN architecture utilizes a neural network backbone to dynamically generate weights for a linear regression output layer. This structure provides a foundation for local interpretability, allowing the final prediction to be decomposed into the individual contributions of each input feature. Furthermore, the local weights can be aggregated across the dataset to yield global feature importance scores~\cite{kadra2024interpretable}, providing both instance-level and dataset-level knowledge. However, the experiments in this paper reveal a critical vulnerability in this approach: the model lacks the necessary constraints to guarantee reliable explanations. Because the highly expressive neural network backbone is unconstrained, it can arbitrarily assign disproportionate magnitude to a single feature's weight while suppressing the others, without altering the final prediction. Consequently, the model can achieve excellent predictive accuracy while yielding explanations that are completely unfaithful. For instance, in our experiments, the model could suppress all feature weights to near zero except for the intercept, leading to the false interpretation that none of the features had an effect on the prediction. which was far from true.

To resolve this fundamental flaw, we introduce in this paper a novel regularization mechanism that forces the generated regression weights to agree with the local variations present in the training data. In high-dimensional feature spaces, data sparsity often makes the reliable computation of these local variations challenging. To address this issue, we employ an interpolation strategy inspired by the Synthetic Minority Over-sampling Technique (SMOTE) to generate localized synthetic samples, effectively densifying the data neighborhoods. This targeted local fidelity regularization (LFR) not only guarantees high-fidelity interpretability but also yields consistent improvements in predictive performance over the original IMN across a diverse suite of real-world tabular datasets from the OpenML benchmark~\cite{bischlopenml}. The predictive performance is competitive with state-of-the-art methods such as CatBoost~\cite{prokhorenkova2018catboost} and TabNet.

Our main contributions are summarized as follows:
\begin{itemize}
    \item We introduce LFR, a novel regularization mechanism for the IMN architecture that ensures reliable local interpretations by aligning regression weights with local data variations.
    \item We empirically demonstrate that this targeted regularization not only resolves interpretability flaws but also consistently enhances predictive performance, yielding results highly competitive with current state-of-the-art black-box models for tabular data.
    \item We provide an open-source, highly modular Python implementation of the LFR-IMN framework, designed with a familiar scikit-learn style API to ensure ease of use. The repository is publicly available at \url{https://github.com/hugohammer/LFR-IMN.git}.
\end{itemize}

\section{Interpretable Mesomorphic Neural Networks}

In this section, we provide a brief description of the IMN model, including how it is trained and interpreted as proposed in \cite{kadra2024interpretable}. Let the data matrix $\mathbf{X} \in \mathbb{R}^{n \times p}$ represent a tabular dataset comprising $n$ data points and $p$ features. Let $\mathbf{x}_i = (x_{i1}, \ldots, x_{ip})$ for $i \in \{1, \dots, n\}$ denote a specific data point (a row of the data matrix), and let $\mathbf{x}^{(j)} = (x_{1j}, \ldots, x_{nj})$ for $j \in \{1, \ldots, p\}$ denote a specific feature (column). Let $\mathbf{y} = (y_1, \ldots, y_n)$ represent the output variable. The outputs can be continuous, $y_i \in \mathbb{R}$ for $i \in \{1, \ldots, n\}$, or categorical over $C$ categories, $y_i \in \{1, \ldots, C\}$.
 
\begin{figure}
\centering
\begin{tikzpicture}[
    >=stealth, 
    thick,
    input node/.style={circle, draw=blue!70, fill=blue!5, minimum size=1.2cm, font=\large},
    weight node/.style={rectangle, draw=green!70!black, fill=green!5, minimum height=0.8cm, minimum width=2.8cm, rounded corners, font=\normalsize},
    output node/.style={circle, draw=red!70, fill=red!5, minimum size=1.2cm, font=\Large},
    nn box/.style={rectangle, draw=orange!80, fill=orange!10, minimum width=3.0cm, minimum height=7cm, rounded corners, align=center, font=\large\bfseries}
]

\node[nn box] (nn) at (3.5, 1) {Neural\\[0.2em]Network\\[0.2em]Backbone};

\node[above=0.5cm of nn, font=\large] (nn_eq) {$g(\mathbf{x}_i, \theta)$};

\node[input node] (x) at (0, 1) {$\mathbf{x}_i$};

\node[font=\bfseries\large] at (x |- nn_eq) {Input Features};

\draw[->] (x) -- (nn.west |- x);

\node[weight node] (w0) at (7.5, 4) {$w_0(\mathbf{x}_i, \theta)$};
\node[weight node] (w1) at (7.5, 2) {$w_1(\mathbf{x}_i, \theta)$};
\node[weight node] (w2) at (7.5, 0) {$w_2(\mathbf{x}_i, \theta)$};
\node[font=\Large] (w_dots) at (7.5, -1) {$\vdots$};
\node[weight node] (wp) at (7.5, -2) {$w_p(\mathbf{x}_i, \theta)$};

\draw[->] (nn.east |- w0) -- (w0.west);
\draw[->] (nn.east |- w1) -- (w1.west);
\draw[->] (nn.east |- w2) -- (w2.west);
\draw[->] (nn.east |- wp) -- (wp.west);

\node[output node] (y) at (11.5, 1) {$\hat{y}_i$};

\draw[->] (w0.east) -- (y);
\draw[->] (w1.east) -- (y);
\draw[->] (w2.east) -- (y);
\draw[->] (wp.east) -- (y);

\draw[->, rounded corners=8pt] (x.south) -- (0, -3.5) -| (y.south);

\node[above=3cm of y, xshift=-1.5cm, font=\large] (out_eq) {$\hat{y}_i = w_0(\mathbf{x}_i,\theta) + \sum_{j=1}^p w_j(\mathbf{x}_i, \theta) x_{ij}$};

\end{tikzpicture}
\caption{The Interpretable Mesomorphic Neural Network (IMN) architecture. The generic neural network backbone $g$ takes a data point as input and dynamically generates a set of linear weights. These weights act as local coefficients to produce the final prediction $\hat{y}_i$.}
\label{fig:imn_architecture}
\end{figure}
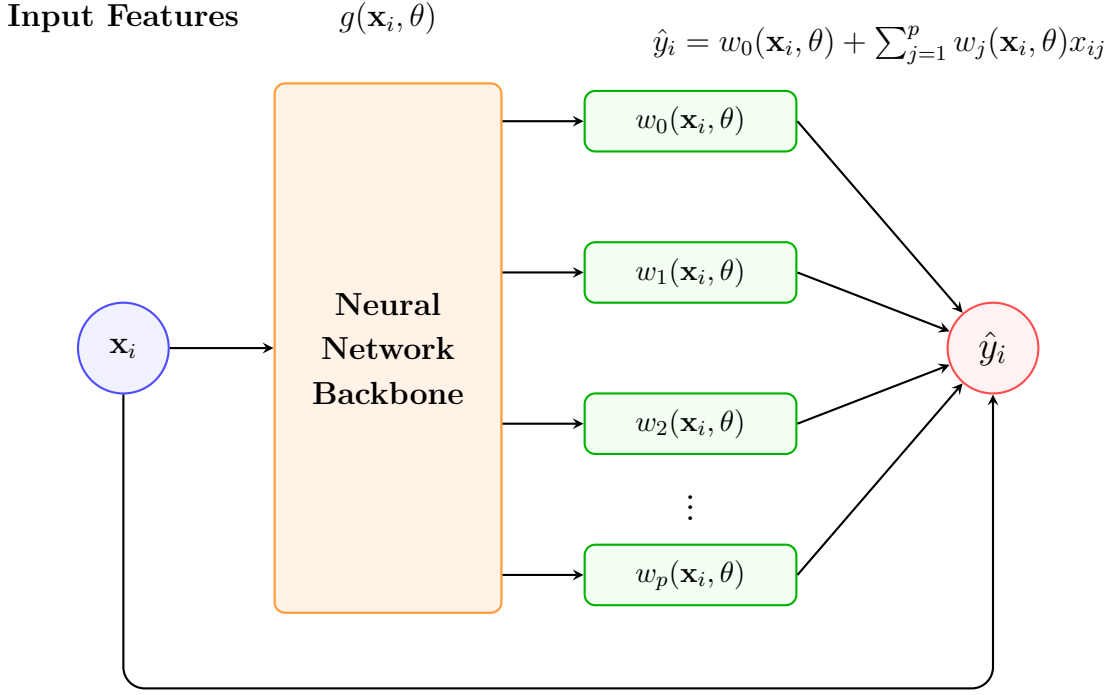

We start by presenting the model when the output is continuous. The architecture comprises of two main parts: The first part is a non-interpretable neural network $g: \mathbb{R}^p \times \Theta \to \mathbb{R}^{p+1}$ taking a data point $\mathbf{x}_i$ as input and returning an $\mathbb{R}^{p+1}$-dimensional vector of weights $\mathbf{w}(\mathbf{x}_i, \theta) = (w_0(\mathbf{x}_i, \theta), \ldots, w_p(\mathbf{x}_i, \theta)) = g(\mathbf{x}_i, \theta)$, where $\theta \in \Theta$ are the parameters of the neural network. The second part is a linear output layer to predict $y_i$ using the generated weights as regression coefficients
\begin{equation}
    \label{eq:1}
    \hat{y}_i = f(\mathbf{x}_i) =  w_0(\mathbf{x}_i,\theta) + \sum_{j=1}^p w_j(\mathbf{x}_i, \theta) x_{ij}
\end{equation}
An illustration of the model is shown in Figure \ref{fig:imn_architecture}.
The model is trained by minimizing a loss function between the data and the model predictions
\begin{equation}
    \hat{\theta} = \argmin_{\theta} \sum_{i=1}^n \left[\mathcal{L}\left(y_i, w_0(\mathbf{x}_i,\theta) + \sum_{j=1}^p w_j(\mathbf{x}_i, \theta) x_{ij}\right) + \lambda \|\mathbf{w}(\mathbf{x}_i, \theta)\|_1 \right]
\end{equation}

To interpret a prediction, the idea is to view the regression layer in Eq. \ref{eq:1} as an ordinary regression model, such that the local coefficient (or effect size) of each feature $j$ is given directly by the weight $w_j(\mathbf{x}_i, \theta)$, while its total contribution to the prediction is $w_j(\mathbf{x}_i, \theta) x_{ij}$.

Alternatively, if the output is categorical over $C$ categories, the architecture is naturally extended to predict class probabilities using a softmax function. In this setting, the neural network backbone is modified to output a distinct set of weights for each class $c \in \{1, \dots, C\}$. Let $\mathbf{w}_c(\mathbf{x}_i, \theta) = (w_{c,0}(\mathbf{x}_i, \theta), \dots, w_{c,p}(\mathbf{x}_i, \theta))$ denote the weight vector associated with class $c$. The logit (pre-activation value) for class $c$, denoted as $z_{ic}$, is computed via the same linear combination:
\begin{equation}
  z_{ic} = w_{c,0}(\mathbf{x}_i, \theta) + \sum_{j=1}^p w_{c,j}(\mathbf{x}_i, \theta) x_{ij}  
\end{equation}
The predicted probability for the data point belonging to class $c$ is then obtained by applying the softmax function across all class logits
\begin{equation}
  \hat{p}_{ic} = \frac{\exp(z_{ic})}{\sum_{k=1}^C \exp(z_{ik})}  
\end{equation}
To interpret the classification prediction, the analysis is performed directly on the logits, $z_{ic}$, rather than the non-linear softmax probabilities. Because the logit computation relies on the exact same linear combination of the input features as the continuous formulation, we effectively treat the logit of each class as an independent ordinary regression model. Consequently, the local contribution of feature $j$ toward predicting class $c$ is directly given by the weight $w_{c,j}(\mathbf{x}_i, \theta)$, allowing the interpretability framework established for the linear regression case to apply seamlessly to classification.

\section{Local Fidelity Regularization}

While the IMN architecture provides a highly flexible framework for generating instance-wise interpretations, the formulation relies on an implicit and precarious assumption: that the neural network will naturally assign weights $w_j(\mathbf{x}_i, \theta)$ that faithfully represent the true local importance of each feature $j$. However, as defined in Eq. \ref{eq:1}, the model only constrains the dot product of the weights and the features to match the target variable. Because the weight-generating function $g(\mathbf{x}_i, \theta)$ is a highly expressive neural network, the model is severely underdetermined. 

To illustrate the severity of this vulnerability, consider a simple, perfectly linear data-generating process where the true target is given by $y = h(\mathbf{x}) = b_1 x^{(1)} + b_2 x^{(2)}$. In an ideal interpretable model, the network should recover the ground truth coefficients, outputting the weights $w_1(\mathbf{x}_i, \theta) = b_1$ and $w_2(\mathbf{x}_i, \theta) = b_2$ (with $w_0 = 0$). However, without strict local constraints, an unregularized IMN can easily discover degenerate solutions. For instance, the network could output $w_1(\mathbf{x}_i, \theta) = (b_1 x_i^{(1)} + b_2 x_i^{(2)})/x_i^{(1)}$ and $w_2(\mathbf{x}_i, \theta) = 0$.  Both weight configurations yield the exact same mathematically optimal prediction $\hat{y}(\mathbf{x}_i) = b_1 x_i^{(1)} + b_2 x_i^{(2)}$. For example, assuming the true effects are $b_1=1$ and $b_2=3$, the degenerate solution at the data point $x^{(1)}=2, x^{(2)}=4$ yields local weights of $w_1 = (1 \cdot 2 + 3 \cdot 4) / 2 = 7$ and $w_2=0$, which drastically misrepresents the true underlying effects ($b1 = 1, b_2 = 3$). Paradoxically, the $L_1$ penalty used in the original IMN architecture actively encourages this degenerate behaviour by heavily penalizing dense weight distributions, thereby pushing the network to collapse the predictive variance into a single, highly incorrect feature weight. In such cases, the model achieves perfect predictive performance while providing completely false interpretations.

To prevent the network from exploiting such degenerate solutions, we must enforce that the generated weights represent the true local relationship between the features and the output. If the weights $w_j(\mathbf{x}_i, \theta)$ are to act as the coefficients of a valid local linear approximation, they should ideally correspond to the partial derivatives of the prediction function with respect to each feature. An ideal regularization would therefore penalize the discrepancy between the generated weight and the model's analytical gradient
\begin{equation}
    \label{eq:ideal_reg}
    \mathcal{L}_{\text{Grad}}(\mathbf{x}_i, \theta) = \sum_{j=1}^p \left( w_j(\mathbf{x}_i, \theta) - \frac{\partial f}{\partial x^{(j)}}(\mathbf{x}_i) \right)^2 = \|\mathbf{w}(\mathbf{x}_i, \theta) - \nabla f(\mathbf{x}_i)\|_2^2,
\end{equation}
For the simple example above, we see that $\frac{\partial f}{\partial x^{(1)}} = b_1$ and $\frac{\partial f}{\partial x^{(2)}} = b_2$, which is substantially different from the weights in the degenerate solution $w_1 = (b_1 x^{(1)} + b_2 x^{(2)})/x^{(1)}$ and $w_2 = 0$. In this scenario, Eq. \ref{eq:ideal_reg} will result in a substantial penalty. In contrast, the weights $w_1 = b_1$ and $w_2 = b_2$ perfectly agree with the partial derivatives and will result in zero penalty. Thus, the regularization pushes the network towards weights that provide reliable model interpretations.

Despite its theoretical appeal, relying on exact partial derivatives presents significant practical disadvantages. First, highly expressive deep neural networks possess decision boundaries that can exhibit rapid, noisy fluctuations at the infinitesimal level. Consequently, the analytical gradient $\frac{\partial f}{\partial x^{(j)}}(\mathbf{x}_i)$ often captures localized noise rather than the meaningful macroscopic trends of the data manifold. Second, from a computational perspective, computing $\mathcal{L}_{\text{Grad}}$ requires calculating the derivative of a derivative (i.e., double backpropagation) with respect to the input features during training. For large tabular datasets with numerous features, this drastically increases memory consumption and computational overhead, rendering the approach inefficient.

To overcome the instability and computational expense of infinitesimal analytical derivatives, we propose \textit{Local Fidelity Regularization} (LFR). Rather than relying on exact partial derivatives, LFR measures the changes in the data over a local neighborhood. The core requirement is that the linear regression model parameterized by the weights at an anchor point $\mathbf{x}_i$ should accurately extrapolate the model's prediction at a nearby point in the feature space, typically referred to as explanation fidelity. 

However, tabular data in high-dimensional spaces is often sparse, meaning naturally occurring nearest neighbors may be too distant to form a reliable local linear neighborhood. To ensure robust and stable estimates of the local variation, we utilize an interpolation strategy inspired by the Synthetic Minority Over-sampling Technique (SMOTE) \cite{chawla2002smote}. 

For a given anchor point $\mathbf{x}_i$, we generate a set of $K$ synthetic neighborhood points. To generate the $k$-th synthetic point $\tilde{\mathbf{x}}_k$, we randomly select a neighboring data point $\mathbf{x}'$ belonging to the same class (or possessing a similar continuous target) and apply linear interpolation:
\begin{equation}
    \tilde{\mathbf{x}}_k = \mathbf{x}_i + \alpha_k (\mathbf{x}' - \mathbf{x}_i),
\end{equation}
where $\alpha_k$ is a scalar uniformly distributed between zero and one. By repeatedly sampling neighbors and $\alpha_k$ values, we can densify the local space with an arbitrary number of synthetic points.

We then enforce local fidelity by requiring the linear approximation anchored at $\mathbf{x}_i$ to match the full neural network's prediction at these synthetic points. Let $\hat{y}_{ext}(\tilde{\mathbf{x}}_k)$ be the prediction at the synthetic point extrapolated purely using the weights generated for the original anchor point $\mathbf{x}_i$:
\begin{equation}
    \hat{y}_{ext}(\tilde{\mathbf{x}}_k) = w_0(\mathbf{x}_i, \theta) + \sum_{j=1}^p w_j(\mathbf{x}_i, \theta) \tilde{x}_{kj}.
\end{equation}
Our proposed LFR penalty is defined as the mean squared difference between this linear extrapolation and the full neural network forward pass, evaluated across the $K$ synthetic points:
\begin{equation}
    \label{eq:3}
    \mathcal{L}_{LFR}(\mathbf{x}_i, \theta) = \frac{1}{K} \sum_{k=1}^K \left( f(\tilde{\mathbf{x}}_k) - \hat{y}_{ext}(\tilde{\mathbf{x}}_k) \right)^2.
\end{equation}
By replacing the computationally expensive analytical gradient with a discrete, SMOTE-stabilized neighborhood penalty, LFR effectively grounds the network's interpretations in the true geometric structure of the data. This computationally efficient constraint successfully guarantees that the generated linear weights reflect realistic local data variations, preventing false feature attributions while simultaneously guiding the network toward robust, generalizable decision boundaries. Combining the penalty with the original IMN, we obtain the local fidelity regularized IMN (LFR-IMN)
\begin{equation}
    \label{eq:2}
    \hat{\theta} = \argmin_{\theta} \sum_{i=1}^n \left[\mathcal{L}\left(y_i, w_0(\mathbf{x}_i,\theta) + \sum_{j=1}^p w_j(\mathbf{x}_i, \theta) x_{ij}\right) + \lambda \|\mathbf{w}(\mathbf{x}_i, \theta)\|_1 + \gamma \mathcal{L}_{LFR}(\mathbf{x}_i, \theta) \right]
\end{equation}
While the proposed LFR mechanism explicitly enforces spatial smoothness and structural fidelity, the $L_1$ regularization ensured improved prediction performance and sparsity and thus was retained. 

\subsection{Tuning the Fidelity Regularization}
\label{sec:tuning}
While the regularization parameter $\lambda$ is selected solely to optimize predictive performance, the parameter $\gamma$ controls the balance between predictive performance and interpretability. Let $\mathcal{I}(\gamma=a)$ denote the measured infidelity of the model on a validation set when the value $a$ is used for $\gamma$ during training (Eq. \ref{eq:2}) and $\lambda$ and other hyperparameters are set to the values that achieve optimal predictive performance on the validation set. The infidelity is computed as the average of $\mathcal{L}_{LFR}(\mathbf{x}_i, \theta)$ in Eq. \ref{eq:3} for all the data points in the validation set. Define
\begin{equation}
    R_{\mathcal{F}} = \frac{\mathcal{I}(\gamma=0)}{\mathcal{I}(\gamma=a) + \mathcal{I}(\gamma=0)}
\end{equation}
which compares the fidelity using $\gamma = a$ during training relative to $\gamma = 0$, where $\gamma = 0$ indicates that no fidelity regularization is used during training (thus reverting to the original IMN model). If $\mathcal{I}(\gamma=a) = 0$, representing perfectly faithful feature attributions, $R_{\mathcal{F}} = 1$. If $\mathcal{I}(\gamma=a) = \mathcal{I}(\gamma=0)$, then $R_{\mathcal{F}} = 1/2$, and if $\mathcal{I}(\gamma=a)$ becomes very large, $R_{\mathcal{F}}$ will approach zero. Let $D \in [0,1]$ represent to a predictive performance metric on the $[0,1]$ interval where a higher score is better (e.g., AUROC or $R^2$). We suggest optimizing the $\gamma$ parameter by maximizing the score
\begin{equation}
    \label{eq:4}
    S(\gamma = a) = \alpha D + (1 - \alpha) R_{\mathcal{F}}
\end{equation}
where $\alpha \in [0,1]$ controls the relative importance assigned to predictive performance and interpretability. Larger values of $\alpha$ place greater emphasis on predictive performance, whereas smaller values assign greater weight to explanation fidelity.

In principle, increasing interpretability may come at the expense of predictive performance, resulting in a trade-off between the two objectives. However, in our experiments, we observed a positive relationship between predictive performance and interpretability. Models with higher fidelity ratios generally achieved higher AUROC values, implying that improved interpretability did not lead to a degradation in predictive performance.

As a result, using $\alpha = 0.5$, striking a fair balance between predictive performance and interpretablity, the optimal score, $S$, was typically obtained for a $\gamma$ that close to maximal AUROC while simultaneously producing fidelity ratios close to one, generally in the range of $0.98-0.99$. Thus, rather than revealing a conflict between predictive performance and interpretability, the proposed score primarily served to identify models that achieved both high predictive performance and faithful explanations.

\section{Synthetic Experiment}

In this section, we conduct a synthetic experiment to evaluate the efficacy of the proposed LFR-IMN model. The primary objective is to compare LFR-IMN against the original IMN across two dimensions: predictive performance and reliability of interpretability. Specifically, we aim to verify whether the generated instance-wise weights $w_j(\mathbf{x}_i, \theta)$ accurately recover the true local effects of the underlying data-generating process, defined by the analytical partial derivatives $\partial h/\partial x^{(j)}(\mathbf{x}_i)$, where $h$ is the true model.

To establish a controlled ground truth, we generated a synthetic dataset comprising $2000$ training observations, $500$ validation observations, and $500$ test observations. The input features were sampled from a standard normal distribution, $\mathbf{x}^{(j)} \sim \mathcal{N}(0, 1)$ for $j \in \{1, 2\}$. The target variable was generated using a continuous smooth function
\begin{equation}
  y = h(\mathbf{x}) + \epsilon = 1.0 + x_1 + x_2 + x_1 x_2 + \epsilon  
\end{equation}
where $\epsilon \sim \mathcal{N}(0, 0.1^2)$ represents Gaussian noise. Consequently, the theoretical optimal test Mean Squared Error (MSE) is bounded by the noise variance at approximately $0.01$. The inclusion of the interaction term $x_1 x_2$ ensures that the true local effects of $x_1$ and $x_2$ change dynamically across the input space. Therefore, a faithful interpretable model must actively adjust its local weights $w_j(\mathbf{x}_i, \theta)$ depending on the specific location of the data point $\mathbf{x}_i$.

To determine the optimal configurations, we performed a hyperparameter sweep over the $L_1$ sparsity penalty $\lambda \in \{0.01, 0.1, 1.0\}$ and the LFR parameter $\gamma \in \{0.01, 0.1, 1.0, 10.0, 100.0\}$. The LFR parameter, $\gamma$, was tuned using $S(\gamma = a)$ in Eq. \ref{eq:4} with $\alpha = 0.5$ and $R^2$ was used as the predictive performance metric $D$. To ensure objectivity in our evaluation, the experiment was executed using an arbitrary random seed, 42, and the local explanations were extracted for an arbitrarily selected test instance, 42. The feature values for this specific instance were $x_1 = -0.0951$ and $x_2 = -0.3330$ and the analytical true gradients were $\partial h/\partial x^{(1)} = 1.0 + x_2 = 0.6670$, $\partial h/\partial x^{(2)} = 1.0 + x_1 = 0.9049$, and the true local intercept was $1.0 - x_1 x_2 = 0.9683$.

The predictive performance and local explainability results for both the MLP and TabResNet backbones are presented in Table \ref{tab:synth_mlp} and Table \ref{tab:synth_tabresnet}, respectively. As shown in Table \ref{tab:synth_mlp}, when utilizing the MLP backbone, the test MSE for LFR-IMN is on par with the original IMN, with both models approaching the theoretical noise floor. However, we observe that the weights generated by the unregularized IMN are completely unreliable. To minimize the $L_1$ penalty, the network ignores the feature interactions entirely, driving the local gradients to near zero and absorbing all predictive magnitude into the intercept. In other words, the interpretation would be that, locally, the feature do not have any effect on the prediction. In stark contrast, LFR-IMN successfully neutralizes this degenerate behaviour, almost perfectly recovering the true local effects while maintaining excellent predictive accuracy.
\begin{table}
\centering
\caption{Predictive performance and interpretability with the MLP backbone. The first row shows MSE on the testset. The second column (True) shows the intercept and the true local effects for the arbitrary data point given by $\partial h/\partial x^{(1)}$ and $\partial h/\partial x^{(2)}$. The two rightmost columns show the weights $w_0(\mathbf{x}_i, \theta)$ to $w_2(\mathbf{x}_i, \theta)$ for IMN and LFR-IMN, respectively.}
\label{tab:synth_mlp}
\begin{tabular}{l|c|c|c}
\hline
\textbf{Metric} & \textbf{True} & \textbf{IMN} ($\lambda=0.01$) & \textbf{LFR-IMN} ($\gamma=100.0, \lambda=0.01$) \\ \hline
Test MSE & 0.0100 & 0.0115 & 0.0118 \\
Intercept & 0.9683 & 0.5930 & 0.9676 \\
Effect $x_1$ & 0.6670 & $-0.0046$ & 0.6562 \\
Effect $x_2$ & 0.9049 & 0.0090 & 0.9130 \\ \hline
\end{tabular}
\end{table}
\begin{table}
\centering
\caption{Predictive performance and interpretability with the TabResNet backbone. The first row shows MSE on the testset. The second column (True) shows the intercept and the true local effects for the arbitrary data point given by $\partial h/\partial x^{(1)}$ and $\partial h/\partial x^{(2)}$. The two rightmost columns show the weights $w_0(\mathbf{x}_i, \theta)$ to $w_2(\mathbf{x}_i, \theta)$ for IMN and LFR-IMN, respectively.}
\label{tab:synth_tabresnet}
\begin{tabular}{l|c|c|c}
\hline
\textbf{Metric} & \textbf{True} & \textbf{IMN} ($\lambda=0.01$) & \textbf{LFR-IMN} ($\gamma=10.0, \lambda=0.01$) \\ \hline
Test MSE & 0.0100 & 0.0444 & 0.0244 \\
Intercept & 0.9683 & 0.8883 & 0.9674 \\
Effect $x_1$ & 0.6670 & 0.4877 & 0.5673 \\
Effect $x_2$ & 0.9049 & 0.7017 & 0.9039 \\ \hline
\end{tabular}
\end{table}

When evaluating the models using the more complex TabResNet backbone (Table \ref{tab:synth_tabresnet}), the advantages of LFR are even more pronounced. First, we observe that both models achieve poorer predictive performance compared to using an MLP backbone. This is expected given the inherent simplicity and smoothness of the underlying data-generating function relative to TabResNet's highly expressive capacity. Furthermore, we observer that LFR-IMN obtains substantially better predictive performance (MSE $= 0.0244$) than the unregularized IMN (MSE $= 0.0444$). At the same time, the LFR-IMN weights are substantially closer to the true analytical gradients. This demonstrates a clear dual benefit: LFR not only enforces structurally reliable interpretations but also substantially improves predictive performance.

\section{XAI Benchmark Evaluation}

To rigorously evaluate the quality of the explanations generated by our proposed method, we utilized the XAI-Bench framework \cite{xai-bench-2021}, which provides synthetic datasets with known ground-truth local feature importances. The XAI-Bench framework samples input features from a multivariate Gaussian distribution and maps them to continuous targets using three increasingly complex functional forms: Gaussian Linear, Gaussian Non-Linear, and Gaussian Piecewise. The Linear dataset models a standard linear combination of the inputs, the Non-Linear dataset applies complex additive transformations, and the Piecewise dataset employs thresholding to create a discontinuous, step-like regression surface. We refer the reader to the original XAI-Bench paper for further details~\cite{xai-bench-2021}.

We compared LFR-IMN against the unregularized IMN and three baseline explanation strategies applied to the non-interpretable TabResNet model: Local Interpretable Model-agnostic Explanations (LIME) \cite{ribeiro2016should}, SHapley Additive exPlanations (SHAP) \cite{lundberg2017unified}, and a random baseline where feature importances were drawn from a standard normal distribution. We evaluated the quality of the explanations for the original IMN and LFR-IMN for several deep learning backbones, namely TabResNet, Multilayer Perceptrons (MLP), FT-Transformer \cite{gorishniy2021revisiting}, and Neural Oblivious Decision Ensembles (NODE) \cite{PopovMB20}.

\subsection{Metrics and Tuning}
We evaluated the quality of explanations using standard XAI evaluation frameworks \cite{hedstrom2023quantus}, including general Faithfulness, Shapley Correlation, and the retraining-based Faithfulness and Monotonicity metrics introduced by the ROAR (Remove And Retrain) protocol \cite{hooker2019benchmark}. While these metrics are standard in the literature, they measure the \textit{ranking} or \textit{order} of feature importance rather than the absolute scale of the local effects. 

To address this, we introduce \textit{Scaled Infidelity} as our primary evaluation metric. Standard infidelity measures the expected squared difference between a model's actual response to a local perturbation and the response predicted by the explanation. However, this raw metric is highly sensitive to the scale of the perturbation itself. By explicitly normalizing this raw infidelity by the variance of the perturbations, Scaled Infidelity effectively becomes the MSE of the local gradients. This provides a strictly mathematically grounded measurement of how accurately the explanations capture the true, absolute scale of the model's local decision surface. 

Similar to the previous example, $\gamma$ was tuned using $S(\gamma)$ in Eq. \ref{eq:4} with $\alpha = 0.5$ and $R^2$ as the predictive performance metric.

\subsection{Benchmark Results and Discussion}

The benchmark results utilizing the TabResNet backbone are presented in Table \ref{tab:main_xai}. For brevity, the evaluations of the MLP, FT-Transformer, and NODE backbones are provided in Table \ref{tab:appendix_xai} in Appendix \ref{sec:appendix_xai}. 

\begin{table}[ht]
\centering
\setlength{\tabcolsep}{4pt}
\caption{XAI Benchmark results using a TabResNet backbone across ten seeds (mean $\pm$ 95\% confidence interval). The best performing method per row is highlighted in bold. The first, second, and third blocks of rows correspond to the Gaussian Linear, Gaussian Non-Linear, and Gaussian Piecewise datasets, respectively.}
\label{tab:main_xai}
\begin{tabular}{l|ccc|cc}
\hline
\textbf{Metric} & \textbf{Random} & \textbf{LIME} & \textbf{SHAP} & \textbf{IMN} & \textbf{LFR-IMN} \\ \hline
Scaled Infidelity ($\downarrow$) & 5.98 $\pm$ 0.20 & 0.00 $\pm$ 0.00 & 0.09 $\pm$ 0.01 & 0.02 $\pm$ 0.06 & \textbf{0.00 $\pm$ 0.00} \\
Faithfulness ($\uparrow$) & 0.58 $\pm$ 0.04 & 0.79 $\pm$ 0.05 & 0.79 $\pm$ 0.05 & 0.78 $\pm$ 0.05 & \textbf{0.80 $\pm$ 0.05} \\
Faith. (ROAR) ($\uparrow$) & 0.67 $\pm$ 0.03 & \textbf{1.00 $\pm$ 0.00} & 1.00 $\pm$ 0.00 & 1.00 $\pm$ 0.00 & 1.00 $\pm$ 0.00 \\
Mono. (ROAR) ($\uparrow$) & \textbf{0.12 $\pm$ 0.01} & 0.08 $\pm$ 0.02 & 0.09 $\pm$ 0.02 & 0.08 $\pm$ 0.02 & 0.10 $\pm$ 0.01 \\
Shapley Corr ($\uparrow$) & 0.07 $\pm$ 0.02 & 0.89 $\pm$ 0.00 & 0.88 $\pm$ 0.00 & 0.88 $\pm$ 0.01 & \textbf{0.89 $\pm$ 0.00} \\ \hline
Scaled Infidelity ($\downarrow$) & 6.00 $\pm$ 0.24 & 0.43 $\pm$ 0.09 & 1.61 $\pm$ 0.35 & 0.30 $\pm$ 0.15 & \textbf{0.09 $\pm$ 0.03} \\
Faithfulness ($\uparrow$) & 0.56 $\pm$ 0.06 & \textbf{0.66 $\pm$ 0.07} & 0.62 $\pm$ 0.08 & 0.61 $\pm$ 0.08 & 0.61 $\pm$ 0.10 \\
Faith. (ROAR) ($\uparrow$) & 1.01 $\pm$ 0.10 & 1.35 $\pm$ 0.08 & 1.26 $\pm$ 0.07 & \textbf{1.38 $\pm$ 0.07} & 1.29 $\pm$ 0.10 \\
Mono. (ROAR) ($\uparrow$) & 0.17 $\pm$ 0.03 & 0.15 $\pm$ 0.04 & \textbf{0.18 $\pm$ 0.04} & 0.14 $\pm$ 0.03 & 0.13 $\pm$ 0.03 \\
Shapley Corr ($\uparrow$) & 0.07 $\pm$ 0.03 & \textbf{0.13 $\pm$ 0.02} & 0.09 $\pm$ 0.03 & -0.03 $\pm$ 0.07 & -0.04 $\pm$ 0.05 \\ \hline
Scaled Infidelity ($\downarrow$) & 8.81 $\pm$ 0.72 & 3.20 $\pm$ 0.70 & 4.38 $\pm$ 0.86 & 3.95 $\pm$ 1.24 & \textbf{0.13 $\pm$ 0.06} \\
Faithfulness ($\uparrow$) & 0.65 $\pm$ 0.03 & 0.79 $\pm$ 0.04 & \textbf{0.86 $\pm$ 0.05} & 0.76 $\pm$ 0.10 & 0.72 $\pm$ 0.04 \\
Faith. (ROAR) ($\uparrow$) & 0.67 $\pm$ 0.07 & 0.92 $\pm$ 0.09 & \textbf{1.10 $\pm$ 0.09} & 0.88 $\pm$ 0.18 & 0.97 $\pm$ 0.02 \\
Mono. (ROAR) ($\uparrow$) & \textbf{0.11 $\pm$ 0.01} & 0.08 $\pm$ 0.01 & 0.09 $\pm$ 0.02 & 0.10 $\pm$ 0.02 & 0.07 $\pm$ 0.02 \\
Shapley Corr ($\uparrow$) & 0.06 $\pm$ 0.03 & \textbf{0.82 $\pm$ 0.01} & 0.67 $\pm$ 0.02 & 0.34 $\pm$ 0.19 & 0.81 $\pm$ 0.02 \\ \hline
\end{tabular}
\end{table}

Comparing the performance across the ranking-based metrics (Faithfulness, Monotonicity, and Shapley Correlation), the models and XAI methods generally perform on the same level, with 95\% confidence intervals largely overlapping. A notable exception occurs on the highly complex Gaussian Piecewise dataset, where the unregularized IMN fails to rank features correctly (Shapley Correlation of 0.34 $\pm$ 0.19) while LFR-IMN successfully restores this ranking performance (0.81 $\pm$ 0.02) to match state-of-the-art post-hoc methods. 

However, the most substantial differences are observed for Scaled Infidelity. Here, LFR-IMN performs significantly better than all other methods. Even LIME, which explicitly computes local linear approximations to the TabResNet model, performs significantly worse than LFR-IMN on the non-linear and piecewise datasets. Inspecting the extended backbone analysis in Table \ref{tab:appendix_xai}, this trend is consistent. LFR-IMN dramatically improves Scaled Infidelity across all backbones compared to the unregularized IMN, notably acting as a stabilizer that prevents unreliable interpretability in highly expressive models like the FT-Transformer.

\section{Real-World Datasets}

In this section, we evaluate the predictive performance and explanation fidelity using datasets from the OpenML benchmarking suite, focusing specifically on challenging datasets where standard machine learning methods exhibit relatively low predictive performance~\cite{bischlopenml}. We omitted the largest datasets from the OpenML benchmark suite, as the cumulative computational cost of rigorously training and tuning all evaluated methods on these datasets proved prohibitive. The properties of the datasets included in the evaluation are summarized in Table \ref{tab:dataset_properties}. Hyperparameter optimization was conducted using the Optuna framework, where each model was allocated a budget of 50 trials to maximize the validation AUROC~\cite{akiba2019optuna}.
\begin{table}
\centering
\small
\caption{Summary of the OpenML tabular datasets used in the experiments, detailing the number of instances ($N$), number of features ($P$), and the primary predictive task.}
\label{tab:dataset_properties}
\begin{tabular}{llcl}
\hline
\textbf{ID}  & \textbf{Instances ($N$)} & \textbf{Features ($P$)} & \textbf{Description} \\ \hline
31  & 1000 & 20 & Classify credit risk as good or bad \\
1067  & 2109 & 21 & Predict software defects based on code features \\
1464  & 748 & 4 & Predict whether an individual donated blood \\
40981  & 690 & 14 & Predict credit card application approval \\
41142  & 414 & 1636 & Part of ChaLearn AutoML Challenge \\
41143  & 2984 & 144 & Part of ChaLearn AutoML Challenge \\ 
41164  & 8237 & 800 & Part of ChaLearn AutoML Challenge \\ \hline
\end{tabular}
\end{table}

\begin{figure}
    \centering
    \includegraphics[width = \textwidth]{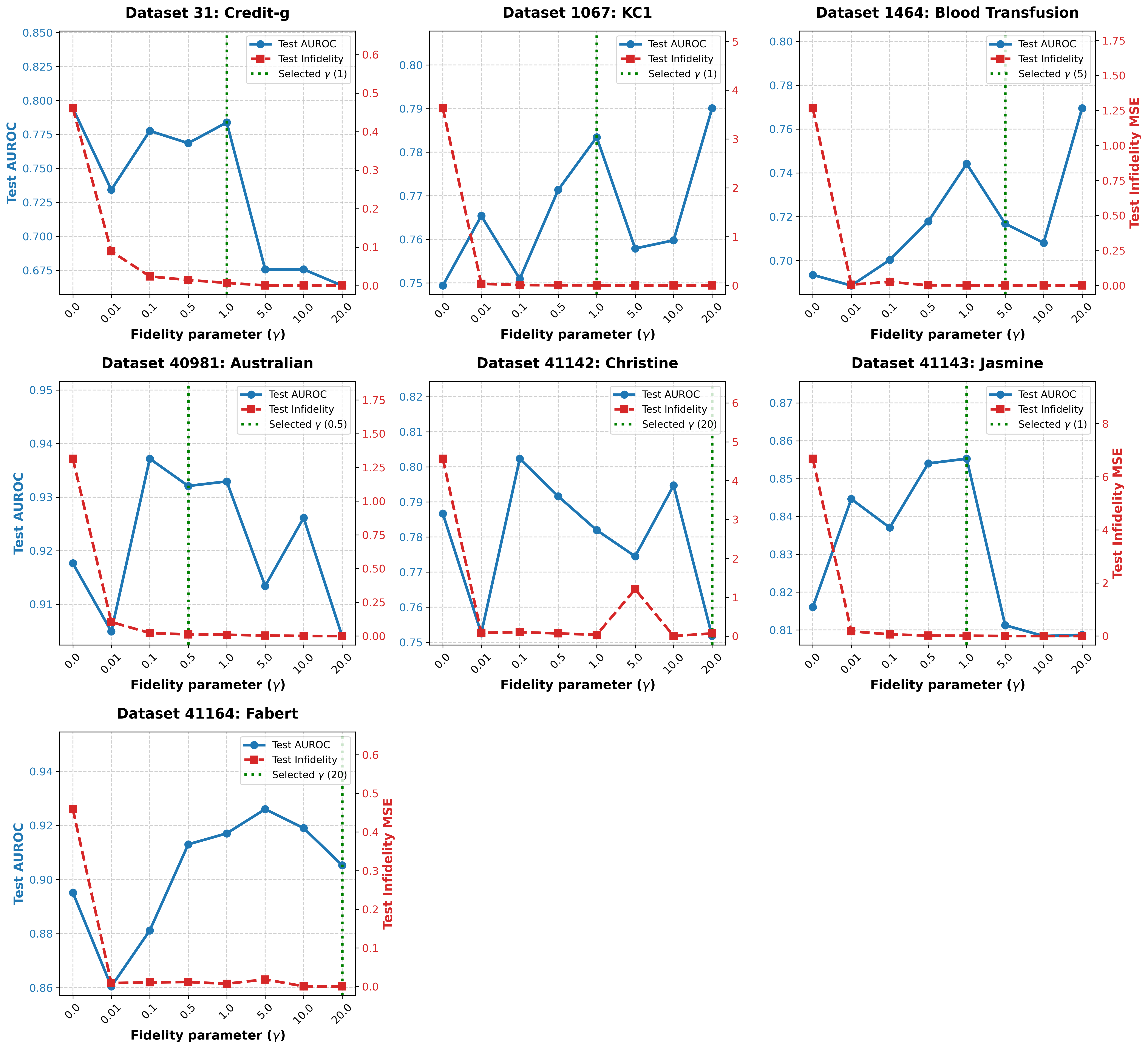}
    \caption{Performance and interpretability across various real-world OpenML benchmark datasets. The blue curves represent predictive performance measured by test AUROC (higher is better). The red squares represent the local explanation error measured by test scaled infidelity MSE (lower is better). The vertical dotted green line shows the value of $\gamma$ selected during the validation phase.}
    \label{fig:tradeoff_realworld}
\end{figure}
Figure \ref{fig:tradeoff_realworld} illustrates both the predictive performance and interpretability across the evaluated datasets. Across all datasets, as the fidelity regularization parameter $\gamma$ increases, the model's infidelity quickly drops to near zero, yielding highly reliable interpretations. Furthermore, increasing $\gamma$ generally has a positive effect on predictive performance. As anticipated, pushing the regularization penalty too far eventually degrades accuracy. Notably, for datasets 41142 and 41164, the validation phase selected $\gamma = 20$, which resulted in noticeably poorer test performance than a smaller penalty (e.g., $\gamma = 0.5$). This discrepancy highlights an occasional high variance in predictive performance, which is particularly pronounced in these high-dimensional datasets (1636 and 800 features, respectively). A more extensive hyperparameter search, such as increasing the number of Optuna trials or ensembling across multiple random seeds, could mitigate this optimization variance, leading to more consistent validation selection and potentially higher test performance for LFR-IMN.

\begin{table}
\centering
\small
\setlength{\tabcolsep}{4pt}
\caption{Predictive performance (AUROC) of various baseline models compared to IMN and LFR-IMN across real-world OpenML datasets. The best performing method for each dataset is highlighted in bold. The bottom row displays the average AUROC across all evaluated datasets.}
\label{tab:baseline_performance}
\begin{tabular}{l|ccccc|cc}
\hline
\textbf{Dataset} & \textbf{LR} & \textbf{RF} & \textbf{CatBoost} & \textbf{TabNet} & \textbf{TabResNet} & \textbf{IMN} & \textbf{LFR-IMN} \\ \hline
31 (Credit-g) & 0.784 & 0.794 & \textbf{0.800} & 0.527 & 0.733 & 0.794 & 0.784 \\
1067 (KC1) & 0.782 & \textbf{0.800} & 0.772 & 0.773 & 0.715 & 0.749 & 0.783 \\
1464 (Blood Transfusion) & \textbf{0.734} & 0.695 & 0.689 & 0.496 & 0.687 & 0.693 & 0.717 \\
40981 (Australian) & 0.915 & 0.921 & \textbf{0.937} & 0.434 & 0.896 & 0.918 & 0.932 \\
41142 (Christine) & 0.787 & 0.808 & \textbf{0.818} & 0.738 & 0.761 & 0.787 & 0.752 \\
41143 (Jasmine) & 0.825 & 0.843 & 0.849 & 0.785 & 0.828 & 0.816 & \textbf{0.855} \\
41164 (Fabert) & 0.913 & 0.933 & \textbf{0.937} & 0.884 & 0.889 & 0.895 & 0.905 \\
\hline
\textbf{Average} & 0.820 & 0.828 & \textbf{0.829} & 0.662 & 0.787 & 0.808 & 0.818 \\ \hline
\end{tabular}
\end{table}
Table \ref{tab:baseline_performance} presents the predictive performance (AUROC) for the benchmark machine learning models Logistic Regression (LR), Random Forest (RF), CatBoost, TabNet, and TabResNet as well as the original IMN and the proposed LFR-IMN. On average, LFR-IMN outperforms the unregularized IMN and achieves results competitive with the best-performing tree-based methods. 

To summarize the results in this section, LFR-IMN not only substantially improves the reliability of the interpretations compared to the original IMN, as demonstrated by the red curves in Figure \ref{fig:tradeoff_realworld}, but it also offers robust predictive accuracy that rivals state-of-the-art black-box models.

\section{Conclusion}

In this work, we explored the Interpretable Mesomorphic Neural Network (IMN), an architecture designed to bridge the gap between the predictive performance of deep learning and the transparency of linear models for tabular data. The core innovation of the IMN lies in its use of a highly expressive neural network backbone to dynamically generate weights for a linear output layer. This structure theoretically allows the model's final prediction to be decomposed into interpretable quantities, using these generated linear weights to directly interpret the instance-wise local effects and feature attributions of the different inputs. The weights can further be aggregated across multiple instances, providing global knowledge about the data and the underlying population.

While this represents a highly appealing architecture that seamlessly combines the predictive power of deep neural networks with intuitive linear interpretation, we identified a critical structural vulnerability: the generated weights of the unregularized IMN are inherently unreliable for interpreting true local effects and attributions. Because the neural network backbone is unconstrained, it is prone to discovering degenerate solutions, e.g. collapsing the predictive magnitude into a single weight, such as the intercept. Consequently, the model can yield fundamentally unfaithful feature attributions despite achieving high predictive accuracy.

To resolve this fundamental flaw, we introduced Local Fidelity Regularization (LFR) and the resulting LFR-IMN architecture. LFR is a computationally efficient penalty designed to align the network's linear output layer with the actual local variations present in the data. 

Through rigorous synthetic experiments and evaluations using the XAI-Bench framework, we demonstrated that LFR successfully prevents degenerate weight collapse, allowing the architecture to accurately recover ground-truth local effects and attributions. LFR-IMN systematically demonstrated better explanation performance than both the unregularized IMN and established post-hoc explanation methods, such as LIME and SHAP, across a variety of complex regression surfaces. 

Our experiments on real-world OpenML datasets revealed that LFR-IMN not only guarantees faithful interpretations but also improves overall predictive performance compared to the unregularized IMN, achieving competitive performance with state-of-the-art black-box methods like CatBoost and TabNet. Thus, LFR-IMN represents a robust and transparent alternative for applications where both predictive performance and interpretability are important.

Looking ahead, several promising directions for future research emerge. Notably, experiments on datasets with massive feature spaces (datasets 41142 and 41164) revealed that validation-based selection of $\gamma$ can be susceptible to high variance. Developing robust tuning methodologies for such regimes—such as multi-seed ensembling or constrained optimization—would further enhance the stability and generalizability of the LFR-IMN architecture. Currently, the generated output weights correspond strictly to individual features, making it difficult to disentangle how the model handles underlying interaction effects. An interesting direction is to develop methodologies that ensure these interaction effects are fairly distributed among the local weights, for example, by integrating principles from cooperative game theory~\cite{lundberg2017unified}. Furthermore, while this work focused on tabular data, the IMN architecture has also demonstrated utility across unstructured data modalities, including images and electrocardiogram signals \cite{kadra2024interpretable,thambawita2026ecg}. Extending the LFR framework to guarantee faithful explanations in these complex domains represents a promising direction for interpretable deep learning.

\bibliographystyle{plain}

\clearpage
\appendix
\section{Extended XAI Benchmark Analysis}
\label{sec:appendix_xai}

Table \ref{tab:appendix_xai} presents the extended XAI-Bench results, demonstrating the effect of Local Fidelity Regularization (LFR) across the MLP, FT-Transformer, and NODE backbones. 

\begin{table}[h]
\centering
\small
\setlength{\tabcolsep}{4pt}
\caption{XAI Benchmark results comparing the unregularized IMN against LFR-IMN across various deep learning backbones (mean $\pm$ 95\% confidence interval). The best performing model per backbone-pair is highlighted in bold for the Scaled Infidelity metric. The first, second, and third blocks of rows correspond to the Gaussian Linear, Gaussian Non-Linear, and Gaussian Piecewise datasets, respectively.}
\label{tab:appendix_xai}
\begin{tabular}{l|cc|cc|cc}
\hline
\multirow{2}{*}{\textbf{Metric}} & \multicolumn{2}{c|}{\textbf{MLP}} & \multicolumn{2}{c|}{\textbf{FT}} & \multicolumn{2}{c}{\textbf{NODE}} \\ 
 & \textbf{IMN} & \textbf{LFR-IMN} & \textbf{IMN} & \textbf{LFR-IMN} & \textbf{IMN} & \textbf{LFR-IMN} \\ \hline
Scaled Infidelity ($\downarrow$) & 0.96 $\pm$ 0.08 & \textbf{0.07 $\pm$ 0.01} & 0.00 $\pm$ 0.00 & \textbf{0.00 $\pm$ 0.00} & 0.03 $\pm$ 0.01 & \textbf{0.02 $\pm$ 0.00} \\
Faithfulness ($\uparrow$) & 0.65 $\pm$ 0.07 & 0.80 $\pm$ 0.05 & 0.82 $\pm$ 0.05 & 0.81 $\pm$ 0.05 & 0.80 $\pm$ 0.05 & 0.80 $\pm$ 0.05 \\
Faith. (ROAR) ($\uparrow$) & 0.76 $\pm$ 0.11 & 1.00 $\pm$ 0.00 & 1.00 $\pm$ 0.00 & 1.00 $\pm$ 0.00 & 1.00 $\pm$ 0.00 & 1.00 $\pm$ 0.00 \\
Mono. (ROAR) ($\uparrow$) & 0.12 $\pm$ 0.02 & 0.07 $\pm$ 0.02 & 0.07 $\pm$ 0.03 & 0.10 $\pm$ 0.02 & 0.08 $\pm$ 0.02 & 0.08 $\pm$ 0.03 \\
Shapley Corr ($\uparrow$) & 0.23 $\pm$ 0.13 & 0.88 $\pm$ 0.00 & 0.89 $\pm$ 0.00 & 0.89 $\pm$ 0.00 & 0.88 $\pm$ 0.00 & 0.88 $\pm$ 0.00 \\ \hline
Scaled Infidelity ($\downarrow$) & 1.15 $\pm$ 0.29 & \textbf{0.37 $\pm$ 0.20} & 0.70 $\pm$ 0.29 & \textbf{0.28 $\pm$ 0.33} & 0.84 $\pm$ 0.33 & \textbf{0.55 $\pm$ 0.29} \\
Faithfulness ($\uparrow$) & 0.78 $\pm$ 0.09 & 0.85 $\pm$ 0.10 & 0.82 $\pm$ 0.10 & 0.78 $\pm$ 0.11 & 0.79 $\pm$ 0.10 & 0.83 $\pm$ 0.09 \\
Faith. (ROAR) ($\uparrow$) & 1.41 $\pm$ 0.13 & 1.56 $\pm$ 0.11 & 1.52 $\pm$ 0.11 & 1.39 $\pm$ 0.12 & 1.45 $\pm$ 0.13 & 1.53 $\pm$ 0.11 \\
Mono. (ROAR) ($\uparrow$) & 0.14 $\pm$ 0.04 & 0.13 $\pm$ 0.04 & 0.14 $\pm$ 0.04 & 0.10 $\pm$ 0.02 & 0.14 $\pm$ 0.04 & 0.13 $\pm$ 0.04 \\
Shapley Corr ($\uparrow$) & 0.06 $\pm$ 0.05 & 0.06 $\pm$ 0.04 & 0.13 $\pm$ 0.14 & 0.08 $\pm$ 0.11 & -0.11 $\pm$ 0.14 & -0.01 $\pm$ 0.06 \\ \hline
Scaled Infidelity ($\downarrow$) & 4.47 $\pm$ 0.26 & \textbf{3.45 $\pm$ 0.58} & 19.11 $\pm$ 5.45 & \textbf{3.61 $\pm$ 3.00} & 3.12 $\pm$ 0.27 & \textbf{2.34 $\pm$ 0.45} \\
Faithfulness ($\uparrow$) & 0.70 $\pm$ 0.06 & 0.68 $\pm$ 0.08 & 0.65 $\pm$ 0.09 & 0.69 $\pm$ 0.05 & 0.66 $\pm$ 0.10 & 0.83 $\pm$ 0.04 \\
Faith. (ROAR) ($\uparrow$) & 0.69 $\pm$ 0.09 & 0.72 $\pm$ 0.14 & 0.64 $\pm$ 0.14 & 0.88 $\pm$ 0.12 & 0.72 $\pm$ 0.19 & 1.08 $\pm$ 0.09 \\
Mono. (ROAR) ($\uparrow$) & 0.08 $\pm$ 0.01 & 0.08 $\pm$ 0.01 & 0.12 $\pm$ 0.02 & 0.09 $\pm$ 0.02 & 0.11 $\pm$ 0.03 & 0.09 $\pm$ 0.02 \\
Shapley Corr ($\uparrow$) & -0.13 $\pm$ 0.08 & -0.02 $\pm$ 0.16 & 0.10 $\pm$ 0.19 & 0.75 $\pm$ 0.08 & -0.08 $\pm$ 0.12 & 0.20 $\pm$ 0.11 \\ \hline
\end{tabular}
\end{table}

\end{document}